\pdfoutput=1

\documentclass[10pt, a4paper]{article}

\usepackage{lrec2026}
\usepackage{booktabs}
\usepackage{multirow}
\usepackage{amsmath}
\usepackage{amssymb}
\usepackage{graphicx} 
\usepackage{url}
\usepackage{float}
\usepackage{makecell}
\usepackage{enumitem}
\usepackage{footmisc}
\usepackage[most]{tcolorbox}
\usepackage{tabularx,ragged2e,booktabs}

\newcommand{\hficon}{\raisebox{-0.2\height}{\includegraphics[height=1em]{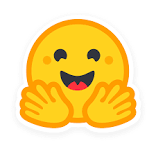}}}
\newcommand{\githubicon}{\raisebox{-0.2\height}{\includegraphics[height=1em]{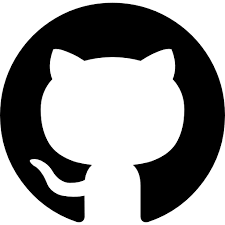}}}

\newcolumntype{Y}{>{\RaggedRight\arraybackslash}X}

\title{LFQA-HP-1M: A Large-Scale Human Preference Dataset for Long-Form Question Answering}

\name{
Rafid Ishrak Jahan \quad
Fahmid Shahriar Iqbal \quad
Sagnik Ray Choudhury
}

\address{
University of North Texas, Denton, Texas, USA \\
\texttt{\{RafidIshrakJahan, FahmidShahriarIqbal\}@my.unt.edu} \\
\texttt{Sagnik.Raychoudhury@unt.edu}
}

\abstract{
Long-form question answering (LFQA) demands nuanced evaluation of multi-sentence explanatory responses, yet existing metrics often fail to reflect human judgment. We present LFQA-HP-1M, a large-scale dataset comprising 1.3M human pairwise preference annotations for LFQA. We propose nine rubrics for answer quality evaluation, and show that simple linear models based on these features perform comparably to state-of-the-art LLM evaluators. We further examine transitivity consistency, positional bias, and verbosity biases in LLM evaluators and demonstrate their vulnerability to adversarial perturbations. Overall, this work provides one of the largest public LFQA preference datasets and a rubric-driven framework for transparent and reliable evaluation.
\\ \newline
} 

\begin{document}
\maketitleabstract


\section{Introduction}

A key strength of modern LLMs is their ability to generate detailed, multi-sentence answers to complex, real-world questions. This field of Long Form Question Answering (LFQA) has advanced substantially over the past years, accompanied by the creation of numerous datasets to evaluate model performance. These include resources where annotators explicitly \textit{write} questions and answers tailored to particular tasks (ShARC \cite{saeidi-etal-2018-interpretation}, NarrativeQA \cite{kovcisky-etal-2018-narrativeqa}, QASPER \cite{dasigi-etal-2021-qasper}), as well as datasets assembled from Q\&A content hosted on public platforms like StackOverflow or Reddit (ELI-5 \cite{fan-etal-2019-eli5}). 
Evaluation on these benchmarks generally relies on reference-based metrics that assess the quality of generated answers by comparing system outputs with one or more human-written reference responses. Such evaluation methods include: a) lexical overlap metrics such as ROUGE \cite{lin-2004-rouge} and BLEU \cite{papineni-etal-2002-bleu} that are insensitive to paraphrasing and deeper semantic equivalence; and b) approaches based on contextual token embeddings such as BERTScore \cite{zhang-etal-2020-bertscore}, as well as learned scoring models like BLEURT \cite{sellam-etal-2020-bleurt} and BARTScore \cite{yuan-etal-2021-bartscore} that estimate semantic similarity. However, these reference-based metrics perform poorly under domain adaptation as they rely on fixed, human-written references and general-domain semantics.

An increasing number of studies focus on \textit{reference-free} evaluation, where proposed methods or models assign direct numerical scores to generated text, without comparing it with reference text. This is commonly known as LLMs-as-judge \cite{li2024llms} when the evaluator models are LLMs. Examples include UniEval \cite{zhong-etal-2022-towards}, GPTScore \cite{fu-etal-2023-gptscore}, and the Prometheus family of models \cite{kim-etal-2024-prometheus}. In many applications, especially LFQA, it is more practical for a model-based evaluator to produce pairwise preference judgments between two candidate answers, as this is generally less demanding. A robust framework for such comparative evaluation helps users make more informed and reliable choices when selecting LLMs for their tasks.

\begin{figure}[!htbp]
    \centering
    \includegraphics[width=\linewidth]{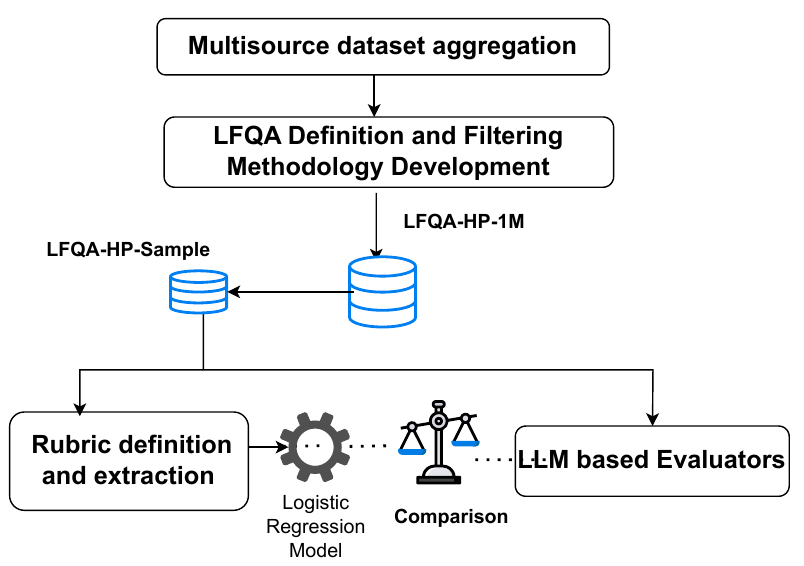}
    \caption{Workflow for our framework: from multisource dataset aggregation and LFQA filtering to rubric-based evaluation and model comparison.}
    \label{fig:arch_overview}
\end{figure}

We develop a framework \footnote{\href{https://github.com/nlpatunt/lfqa-eval-lrec-2026}{\githubicon\
 https://github.com/nlpatunt/lfqa-eval-lrec-2026}} (Fig. \ref{fig:arch_overview}) to advance the field of comparative reference-free evaluation of LFQA. Our contributions are as follows:

\begin{itemize}[leftmargin=*, itemsep=0pt]
\item \textbf{LFQA-HP-1M: A large-scale preference judgment dataset \footnote{\href{https://huggingface.co/datasets/nlpatunt/LFQA-HP-1M}{\hficon\ https://huggingface.co/datasets/nlpatunt/LFQA-HP-1M}} for LFQA.} We construct a large-scale LFQA dataset comprising 1.3 million human preference annotations, where each question is paired with a) two candidate answers and b) a human judgment indicating which answer is preferred. The dataset is curated from existing large-scale preference judgment datasets available on the web: SHP-2 \cite{pmlr-v162-ethayarajh22a}, LFQA Eval \cite{xu-etal-2023-critical}, and ChatArena \cite{zheng2023judging}. These datasets contain human annotations comparing the quality of multiple long responses to a given prompt. However, not all of these prompts can be classified as questions. To address this, we first develop a rigorous definition of what constitutes a long-form question and operationalize it to construct a high-precision prompt-based filtering pipeline that systematically removes non-long-form question (non-LFQ) utterances (\S \ref{sec:dataset-development}).

\item \textbf{Rubric development for answer quality evaluation.} We propose rubrics for long-form answer evaluation to address the limitations of traditional lexical overlap and semantic metrics. These rubrics cover nine fine-grained aspects, including completeness, accuracy, grammar, and factuality (\S \ref{sec:rubric-development}).

\item \textbf{Model evaluation \& discussions.} We benchmark state-of-the-art closed-source LLMs' ability to mimic human preference judgment on a carefully sampled subset of our data. We compare that with interpretable linear models that use the rubrics as features. We also include discussions about multidimensional adversarial robustness for the LLMs. Specifically, we test for models' a) transitive consistency and propensity toward verbosity and position bias in LFQA preference evaluation, and b) robustness for semantics preserving adversarial perturbations to the answers. At each stage of the framework, multiple challenges are systematically addressed to maintain an optimal balance among model performance, computational efficiency, and overall cost (\S \ref{sec:model_evaluation_discussion}). 
\end{itemize}

Our framework unifies large-scale human preference data, interpretable rubric-based scoring, and systematic robustness analyses, enabling transparent, reproducible, and theoretically grounded reference-free evaluation in LFQA. We publicly release the LFQA-HP-1M dataset and all of our code
to facilitate future work into LFQA evaluation.


\section{Related Work}

LFQA progress has been hindered by limited dataset availability. \citet{fan-etal-2019-eli5} introduced ELI5, a 270K-pair dataset of open-ended ``wh'' and ``how'' questions from Reddit's ExplainLikeIamFive community, establishing LFQA as a distinct research area. In parallel, \citet{nakano2022webgpt} developed WebGPT, a dataset of factually grounded long answers generated via web search and reinforcement learning from human feedback, but its questions remained largely ELI5-style, limiting domain diversity.

Beyond dataset scarcity, prior LFQA studies have offered inconsistent definitions of the task, which our work addresses. \citet{fan-etal-2019-eli5} described LFQA as producing elaborate, in-depth answers to open-ended questions, while \citet{krishna-etal-2021-hurdles} defined it as retrieving and generating paragraph-length responses. \citet{su-etal-2022-read} similarly emphasized generating detailed explanations, whereas \citet{stelmakh-etal-2022-asqa} viewed LFQA as answering questions that require such explanations. \citet{xu-etal-2023-critical} highlighted the generation of long, complex answers using LLMs, and \citet{sachdeva-etal-2025-localizing} framed LFQA as requiring thorough responses to complex queries. These variations reveal the lack of a unified definition, complicating automated LFQA identification. 

In addition to limited datasets and inconsistent definitions, LFQA evaluation remains challenging. \citet{rosset-etal-2021-axiomatic} offered one of the first formalizations of human preference judgment in this setting, emphasizing mathematical consistency (e.g., transitivity, monotonicity) in pairwise comparisons. However, their framework prioritized abstract consistency over rubric-level interpretability, offering little guidance for selecting one answer over another. Other early studies relied on lexical metrics such as ROUGE, BERTScore \citep{zhang-etal-2020-bertscore}, and BARTScore \citep{yuan-etal-2021-bartscore}, yet \citet{xu-etal-2023-critical} showed that these metrics correlate weakly with human judgments. Their work provided the first comprehensive empirical comparison of automatic metrics against expert preferences, revealing that standard measures fail to capture human notions of quality. Although they introduced a multi-dimensional evaluation framework informed by expert feedback, the study did not propose a new scoring system or practical evaluation tool.

Recent studies have explored using LLMs to evaluate long-form answers. \citet{dsouza-etal-2025-yescieval} proposed YESciEVAL, a framework for scientific question answering that decomposes responses into multiple reasoning dimensions and leverages LLMs for multi-aspect scoring aligned with expert annotations. However, its scope is confined to scientific domains, limiting its generalizability to open-domain LFQA. Prometheus \citep{kim-etal-2024-prometheus} extends this idea to open-ended text generation, using fine-tuning on large human-preference datasets to achieve strong agreement with human judgments across multiple evaluation metrics. Focusing specifically on LLM-as-a-judge paradigms, JudgeBench \citep{tan-etal-2024-judgebench} introduced the first standardized benchmark to assess LLM judgment quality, testing for consistency, bias, and alignment with human preferences. It also identified systematic biases such as length and positional bias, though it primarily targets bias detection rather than intrinsic answer quality assessment.

 Collectively, prior research underscores three enduring challenges in LFQA research: the scarcity of large-scale human preference datasets; the poor interpretability of existing automatic evaluation metrics; and the lack of understanding of the factors that drive human preference judgments. This paper addresses these gaps.


\section{Preference Judgment Dataset Development}
\label{sec:dataset-development}
Constructing a large-scale, robust, and reliable dataset for LFQA demands a careful balance between domain diversity, annotation reliability, and scalability. This section explains the dataset-building process, the refinement of LFQA definitions, and the development of cost-effective LFQA filtering methods.

\subsection{Multi-Source Dataset Aggregation}

We initiated data construction by aggregating human preference judgment data from three heterogeneous sources, including both conversational and knowledge-intensive sources: 

\begin{enumerate}[itemsep=0pt, leftmargin=*]
    \item The \textbf{Chatbot Arena Conversation Dataset} consists of 33K crowdsourced conversations with pairwise human preference annotations and 3.3K expert-level human annotations. It contains a wide range of conversational topics and model-generated answers, making it a rich source for further analysis to capture human preference from LLM-generated responses.
    \item The \textbf{LFQA Eval: LFQA Evaluation Dataset}. In contrast to typical conversational datasets, the LFQA\_Eval dataset focuses on questions that require reasoning and detailed, multi-sentence responses. It contains approximately 17K long-form questions with pairwise human preference.
    \item The large-scale \textbf{SHP-2: Stanford Human Preferences Dataset v2} is collected from community-driven discussion portals such as Reddit and Stack Exchange. The dataset was primarily built on the assumption that if comment A was written after comment B but still has a higher score, then A is ostensibly preferred to B. The dataset contains approximately 900K prompts and 4.8M responses, and pairwise human judgments across 129 subject areas. 

\end{enumerate}

\subsection{LFQA Definition} 
\label{subsec:lfqa-def}
Both SHP-2 and ChatBot Arena contain prompts that can not necessarily be characterized as long-form questions. To filter them out, we need to rigorously define long-form questions. The first challenge is to differentiate a question from other types of text utterances, as the syntax alone is insufficient. At an abstract level, a question is an utterance whose illocutionary force is to elicit a response that fills an information gap in the speaker's mental state, and is not necessarily characterized by computational cues such as starting with a ``wh'' word or an overt interrogative intent (``could you please explain..''). For example, utterances such as ``it would help to know your opinion on X'' are valid questions that are not represented in most general question answering (extractive or MCQ) datasets.
 
One can distinguish between a short and a long-form question based on the length of a possible answer, but that alone cannot be the key determining factor, as a question can admit ``correct'' answers of varying lengths. We propose a rigorous and operational definition of long-form questions (LFQ) developed through an iterative annotation process involving five annotators excluding the authors: two Ph.D. students, two master's students, and one undergraduate CS student with high English proficiency.

\newtcolorbox{definitionbox}[2][]{
  colback=gray!5!white,
  colframe=black!75,
  fonttitle=\bfseries,
  title=#2,
  #1,
  floatplacement=!htbp
}

\begin{definitionbox}{LFQA Definition}\label{def:lfqa}
A LFQA must express a single, well-defined information need that requires a detailed, multi-sentence answer involving exposition, explanation, reasoning, exploration, or description of the process. These questions are usually complex or open-ended, often asking ``why'' or ``how'' about a process, reason, or concept in an objective manner, which means answers should be grounded in facts, reasoning, or conceptual explanation.

It must not:
\begin{itemize}
    \item Combine multiple distinct or loosely related sub-questions  
    \item  Request personal advice, express individual preferences, seek personal opinions, and recommendations (e.g., product, place, city, food, restaurant, flight, school, course)
    \item Be answerable with a single fact, a yes/no response, only code generation, direct formula calculation, or creative content (e.g., poem, joke, story, music, song lyrics, image, or audio)  
\end{itemize}
\end{definitionbox}

In the first round, 100 questions from the \textbf{Chatbot Arena Conversation Dataset} were annotated as LFQ or not, with labels based on the majority vote. Inter-annotator agreement, calculated using Gwet's AC1 statistic \footnote{It has been shown to provide more stable estimates than Cohen's/Fleiss' $\kappa$ under class imbalance \cite{gwet2008ac1}}, was 0.65, indicating substantial agreement. Then the LFQA definition (\ref{def:lfqa}) was refined  (see Definition \ref{def:lfqa}) using annotator feedback (see Appendix~\ref{def:appendix_lfqa_initial} for the initial LFQA definition). We conducted another annotation round with the updated definition using longer questions from \textbf{SHP-2} with a Gwet's AC1 score of 0.55, indicating moderate agreement. This implies that the annotators are consistent in identifying LFQA, even under stress testing with challenging and borderline examples.

\subsection{Definition to Prompt-based Filtering}
To evaluate the effectiveness of LFQA definitions, we designed two prompts: one with the initial definition and another with the refined definition. The performance of these prompts was tested using three state-of-the-art LLMs: GPT-4o, Llama-4, and Gemini-2.5, and evaluated using standard metrics, including precision, recall, and F1-score. The ground truth was defined as the majority vote from human annotations. 

As shown in Table~\ref{tab:llm_eval_round1_2}, the prompt derived from the updated definition improved performance across all evaluation metrics. Therefore, we selected the updated prompt for large-scale data filtering. 

\begin{table}[!htbp]
\centering
\resizebox{\linewidth}{!}{%
\begin{tabular}{llccc}
\toprule
\textbf{Sample Set} & \textbf{Model} & \textbf{Precision} & \textbf{Recall} & \textbf{F1 Score} \\
\midrule
\multirow{3}{*}{\makecell[l]{First 100 \\ initial definition}} 
    & llama-4        & {0.7500} & 0.2553 & 0.3810 \\
    & gpt-4o         & 0.9091 & \underline{0.2128} & \underline{0.3448} \\
    & gemini-2.5 & 0.8750 & 0.2979 & 0.4444 \\
\midrule
\multirow{3}{*}{\makecell[l]{Second 100 \\ initial definition}}
    & llama-4        & \underline{0.6905} & 0.6558 & 0.6528\\
    & gpt-4o         & {0.7133} & {0.6485} & {0.6382} \\
    & gemini-2.5     & 0.7313 & \textbf{0.7346} & 0.7293 \\
\midrule
\multirow{3}{*}{\makecell[l]{Second 100 \\ updated definition}}
    & llama-4        & 0.9167 & 0.5893 & 0.7174 \\
    & gpt-4o         & \textbf{0.9459} & {0.6250} & \textbf{0.7527} \\
    & gemini-2.5     & 0.8140 & {0.6250} & 0.7071 \\
\bottomrule
\end{tabular}}
\caption{LLMs evaluation metrics (Precision, Recall, F1) on the first and second 100 samples, with both initial and updated definition.}
\label{tab:llm_eval_round1_2}
\end{table}

\subsection{Scaling}

While GPT-4o achieved the highest precision, recall, and F1 scores with the updated definition prompt, its performance gains over Llama-4 were marginal. Given the dataset's size ($\approx$ 900K data points), cost was a critical consideration. As we use Openrouter APIs \cite{openrouter-2024}, Llama-4 Scout costs \$0.08 per million input tokens and \$0.30 per million output tokens, compared to GPT-4o's \$5.0 and \$15.0, respectively, about 60× higher for inputs and 50× for outputs. Thus, at this scale, the cost difference between the two models becomes substantial, making Llama-4 a more practical choice despite the slight performance trade-off.

\begin{figure}[h!]
    \centering
    \includegraphics[width=\linewidth]{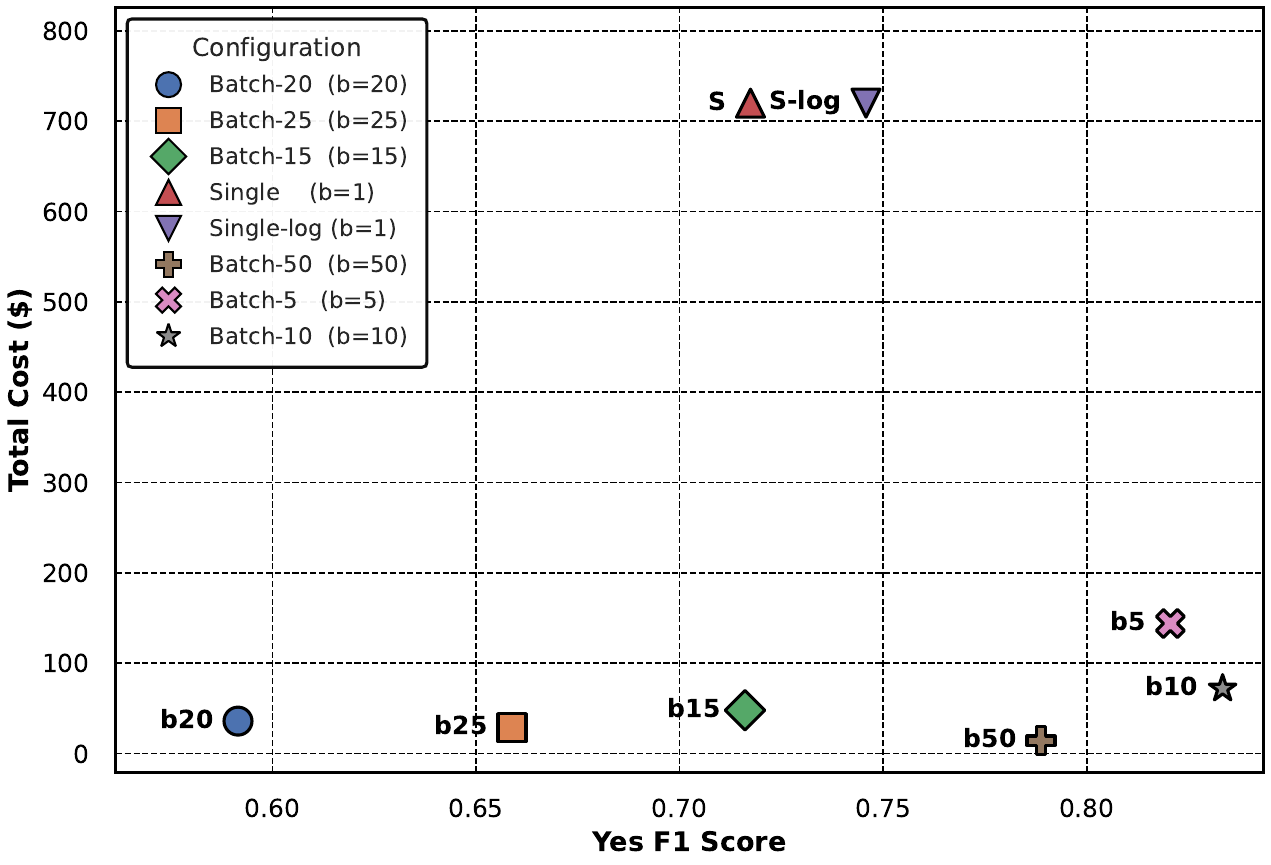}
    \caption{Effect of batch size on filtering performance for Llama-4. Batch sizes above 10 provide marginal cost gains but a large performance drop.}
    \label{fig:yesf1_cost_tradeoff}
\end{figure}

However, even with Llama-4, total processing costs remained substantial. To mitigate this, we first experimented with batching multiple data points but observed performance degradation as the batch size increased, an LLM limitation noted in prior work \cite{batchprompt}. We examined whether the model's internal confidence, as reflected by the \verb|log_score| probability, could serve as a filtering criterion. When we retained only predictions with \verb|log_score| = 0 (i.e., 100\% confidence), 78\% of our test samples met the condition, yielding an F1 score of 0.7174. These results indicated that incorporating \verb|log_score| filtering can improve overall performance. We further explored combining batching with \verb|log_score| filtering to assess the feasibility of cost-efficient processing. In Figure \ref{fig:yesf1_cost_tradeoff}, we plot the F1-score for the positive class against cost. It can be seen that Llama-4, when batching 10 data points and considering only the highest-confidence judgment, achieves the highest performance while maintaining low cost. 

We further evaluated our filtering methodology on the ELI5 subreddit dataset, which is considered one of the earliest large-scale LFQA resources \citep{fan-etal-2019-eli5}. The SHP-2 version of ELI5 contains 24,331 questions. Applying our method, 20,048 questions yielded a \verb|log_score| of 0, among which 19,224 were identified as LFQA. Considering only the model's 100\% confidence responses, the methodology achieved 95.89\% accuracy on the ELI5 dataset.

Our final multi-source dataset comprised approximately 900K unique prompts (from SHP-2, ChatbotArena, and LFQA Eval). After processing them with Llama-4 using a batch size of 10 and mapping back to their corresponding answer pairs, we obtained a 1.3M-instance LFQA pairwise-judgment dataset (LFQA-HP-1M) consisting of 300K unique questions.


\section{Answer Quality Evaluation Rubrics}
\label{sec:rubric-development}

Evaluating LFQA is challenging without standardized metrics. LLM-as-a-judge methods are promising but lack interpretability. This section examines whether a rubric-based evaluation aligns well with human preferences in LFQA.

\subsection{Tools Selection for Rubric Extraction}

We constructed a rubric-based evaluation schema grounded in prior LFQA and text-evaluation research \citep{xu-etal-2023-critical, krishna-etal-2021-hurdles, xian-etal-2025-empirical, hashemi-etal-2024-llm, li-et-al-2025-researchqa} that emphasizes multi-faced evaluation of LFQA. Building on these insights, we employed nine fine-grained rubrics to evaluate the LFQA (see Tab. \ref{tab:rubric_definitions}), ensuring a comprehensive, explainable understanding of answer quality. 

\begin{table*}[!htbp]
\centering
\begin{tabularx}{\textwidth}{lXc}
\toprule
\textbf{Rubrics} & \textbf{Definition} & \textbf{Score range} \\
\midrule
Specificity & measures the precision and depth in explanation over general statements. &  0/1/2, CT \\

Grammar & examines grammatical correctness, adhering to the formal writing style. & 1-5, CO \\

Fluency & measures how clearly the answer communicates. & 1-3, Co \\

Completeness & checks whether the answer addresses all the queries asked by the question. & 1-5, CT \\

Coherence & captures how logically and cohesively the answer is organized. & 1-5, CO \\

Relevance & evaluates whether the answer directly addresses the given question. & 1-5, CO \\

Conciseness & measures how succinct yet sufficiently informative an answer is without redundancy. & 0-1, CO \\

Use of example & checks the presence of a concrete example supporting a relevant explanation. & 0/1, CT \\
    
Factuality & checks claims made in the answer are accurate, verifiable, and free from factual errors or contradictions. & 0-1, CO \\

\bottomrule
\end{tabularx}
\caption{LFQA evaluation rubric definitions and score ranges. CT/CO denotes whether scores are categorical or continuous.}
\label{tab:rubric_definitions}
\end{table*}

\subsubsection{Rubric Value Computation}
The rubrics were designed so that most could be computed using established, non-LLM-based techniques. However, for dimensions such as relevance, coherence, and fluency, the use of LLM-based evaluators is widely accepted in the summarization literature. To operationalize these aspects, we employ G-EVAL, a widely used LLM-driven evaluation framework that relies on carefully constructed prompts to generate reliable scores for these metrics \cite{liu2023geval}. However, the original G-EVAL paper did not provide results for the completeness rubric. We created a G-EVAL style prompt to measure completeness on the YESciEval \cite{dsouza-etal-2025-yescieval} benchmark. The prompt achieved approximately 70\% accuracy on the BioASQ subset, validating its reliability and justifying its use in our experimental setup. For factuality, we leveraged the Veriscore framework, which evaluates the accuracy of verifiable claims in the long text generation \cite{song2023veriscore}. We also optimized the Veriscore framework by replacing GPT-4o with GPT-4o nano, as both model exhibits a high degree of consistency.
    
Specificity was quantified using the SubjECTiveQA-SPECIFIC model by \citet{pardawala2025subjectiveQA}, which offers a generalizable framework for analyzing question–answer pairs using six-dimensional features for measuring subjectivity and informational specificity. 
Conciseness was computed via a lexical density–based metric implemented with the spaCy model ``en\_core\_web\_sm'' \cite{honnibal2020spacy}. Grammar was evaluated using LanguageTool \cite{naber2003languagetool}. Finally, the use of examples was detected through simple prompting to Llama-4, whose results aligned with human judgment. Together, these tools provide a multi-aspect evaluation pipeline to examine LFQA. 

\subsubsection{Score Normalization and Rescaling}
Each selected tool outputs a score corresponding to its designated evaluation rubric (Tab. \ref{tab:rubric_definitions}). Continuous scores that are not between 1-5 (e.g., Conciseness) are rescaled to a 1–5 range using min–max normalization. The use of examples is a binary indicator: 0 = absent, 1 = present. Similarly, the SubjectiveQA-SPECIFIC model outputs three ordinal specificity labels (Negative, Neutral, Positive). Following standard psychometric conventions \citep{stevens1946measurement, bock1972estimating, mccullagh1980regression}, these ordinal categories are treated as partitions of a continuous latent specificity variable, and the logits are converted into a continuous score, which is then rescaled via min–max normalization to a 1–5 range.

\subsection{Rubrics Predict Human Preference}
We trained a binary logistic regression model on pairwise judgment data to identify which rubrics most strongly influence human preference decisions. Each question includes two candidate answers with their corresponding rubric scores, allowing us to examine the predictive power of inter-answer differences across rubrics. To construct a balanced evaluation set, we randomly sampled 5,074 instances from the full dataset, which we refer to as \verb|LFQA-HP-Sample|. The sample was designed to ensure broad coverage across domains, data sources, and annotation types, comprising 750 instances from Chatbot Arena, 750 from LFQA-Eval (balanced between expert and non-expert judgments), 2,039 from SHP-2 Reddit, and 1,536 from SHP-2 StackExchange. The dataset spans 129 distinct domains, with at least 20 questions per domain where available. It was partitioned into training (70\%), development (15\%), and test (15\%) subsets while preserving proportional representation of sources, domains, and annotator expertise. This curated dataset serves as the foundation for rubric-level feature analysis, logistic regression modeling, and benchmarking the performance and robustness of state-of-the-art LLM evaluators.

\begin{table}[!htbp]
\centering
\footnotesize
\begin{tabular}{l r}
\toprule
\textbf{Feature (Rubric Aspect)} & \textbf{Weight} \\
\midrule
Specificity & 0.0420 \\
Grammar & -0.0378 \\
Fluency & 0.0919 \\
Completeness & 0.4065 \\
Coherence & 0.5910 \\
Relevance & -0.3082 \\
Conciseness & 0.1747 \\
Use of Examples & 0.1208 \\
Factuality & 0.0524 \\
\bottomrule
\end{tabular}
\caption{Learned LR weights indicate contribution toward human preference}
\label{tab:logistic_weights}
\end{table}

The LR model weights are presented in Table~\ref{tab:logistic_weights} indicate that a) coherence and completeness are the strongest positive indicators for human preference; b) conciseness, use of example, fluency, factuality, and specificity have a moderate influence.
Interestingly, we observed two negative weights for relevance and grammar rubrics, suggesting that these two features may be constant across both candidate answers. Also, relevance received the highest negative weights, suggesting that overly detailed and verbose answers may receive a high relevance score but were ultimately not preferred by humans.

Overall, the proposed rubrics for long-form answer evaluation provide a comprehensive foundation for quantitative evaluation. 


\section{Model Evaluation and Discussions }
\label{sec:model_evaluation_discussion}

We assess whether state-of-the-art LLMs, in a zero-shot setting, outperform rubric-based linear models at identifying human preferences in LFQA. Specifically, we use three SOTA LLMs: GPT-4o, Gemini-2.5-flash, and Llama-4-Scout. The models are given a question and two answers, and are prompted with instructions to determine the \textit{better} (i.e., the one with higher human preference) one. All models are accessed via their Openrouter APIs \footnote{\url{https://openrouter.ai/}} using the default parameters with a temperature of 0. Each model is prompted three times per data point (a zero temperature does not necessarily guarantee deterministic output \cite{subonis2024zerotemp, he2025nondeterminism}), and the final judgement was determined by the majority response.

Then we examine verbosity and position bias \cite{li2024llms} in the LLMs, i.e., whether the LLM performance is influenced by a) length of the answers rather than the quality of the content, and b) position of the answers, i.e., whether they appear first or second after the question. An assessment of transitive preference judgment is also conducted to evaluate logical consistency and identify potential intransitive cycles. Finally, we evaluate the robustness of LLMs on an adversarially perturbed version of the test dataset, in which each instance included semantically preserving modifications \cite{roth2024token} known to affect the predictions of state-of-the-art Small Language Models such as ModernBERT \cite{warner2024smarter}.   

Table \ref{tab:llm_eval_results} shows that the logistic regression model achieves an accuracy and F1-score of 0.68, comparable to the Gemini-2.5 model (0.68) and Llama-4 model (0.68), and only marginally below the best performing GPT-4o (0.69). This result indicates that interpretable linear classifier models, built solely on fine-grained rubrics, can show comparable performance with SOTA LLMs. The top-performing GPT-4o model was also prompted with detailed definitions for the nine rubrics, but that yielded only a minimal performance improvement. Finally, we fine-tuned the ModernBERT-base model on the \verb|LFQA-HP-Sample| training data, but its performance remained below that of SOTA LLMs. Overall, the tested models achieved below 70\% accuracy, a result slightly higher than the findings of \citet{chen2024preference}, who demonstrated that open-access preference-tuned LLMs consistently struggle to exceed 60\% ranking accuracy on standard preference benchmarks.

\begin{table}[h]
\centering
\scriptsize
\setlength{\tabcolsep}{5pt}  
\begin{tabular}{lcccc}
\toprule
\textbf{Model} & \textbf{Acc.} & \textbf{Prec.} & \textbf{Rec.} & \textbf{F1} \\
\midrule

Logistic Regression & 0.6826 & 0.6821 & 0.6816 & 0.6817 \\
\textit{LLM-as-a-judge} \\
GPT-4o & \textbf{0.6943} & \textbf{0.6970} & \textbf{0.6960} & \textbf{0.6942} \\
Gemini-2.5-flash & 0.6826 & 0.6864 & 0.6847 & 0.6822 \\
Llama-4-Scout & 0.6840 & 0.6899 & 0.6868 & 0.6833 \\

\midrule
\textit{Rubrics} \\
GPT-4o (R) & 0.6979 & 0.6981 & 0.6984 & 0.6979 \\
\midrule
\textit{Fined-Tuned} \\
ModernBERT-base & 0.6635 & 0.6546 & 0.6357 & 0.6450 \\
\bottomrule
\end{tabular}
\caption{Comparative performance of LLM-based evaluators and logistic regression models}
\label{tab:llm_eval_results}
\end{table}

\subsection{Preference Transitivity}
To evaluate ranking consistency, we assess the transitivity of preference judgments. A preference relation is transitive if there exist at least three distinct answers for a unique question, and follows the relation that if $A \succ B$ and $B \succ C$, then $A \succ C$. In contrast, a violation occurs when a 3-cycle $(A \succ B,\; B \succ C,\; C \succ A)$ forms, indicating intransitivity. However, the existence of three answers is necessary but not sufficient for running transitivity violation detection tests (see Table~\ref{tab:transitivity_cases}). 

\begin{table}[h]
\centering
\small
\begin{tabularx}{\linewidth}{lXXX}
\hline
\textbf{Choices} 
& \textbf{Transitive}
& \textbf{Non-Transitive}
& \textbf{Not Testable} \\

\hline
A $\mid$ B & A & A & B \\
B $\mid$ C & B & B & B \\
C $\mid$ A & A & C & -- \\
\hline
\end{tabularx}
\caption{Illustration of transitive, cyclic (3-cycle), and non-testable preference cases. In the final column, the model selects option B in two cases, thereby rendering the third option irrelevant for transitivity detection.}
\label{tab:transitivity_cases}
\end{table}

Among the tested LLM-based judges, GPT-4o achieves a transitivity violation rate of 4.7\%, indicating strong internal consistency. In contrast, Llama-4 achieves a high violation rate of 12.2\%, and Gemini-2.5 exhibits the highest rate of 15.3\%, indicating inconsistencies in judging LFQA preferences.

\subsection{Position and Verbosity Bias}

Biases in model evaluation threaten the reliability of the automated LFQA judging system. As our \verb|LFQA-HP-Sample| dataset consists of pairwise human preferences, we first test whether swapping the answer position or order (e.g., Answer 2 vs Answer 1) changes LLMs' decision in judging pairwise LFQA. Ideally, the standard performance metrics should remain unchanged as the content of each answer remains the same. However, we observed performance dropped in the table \ref{tab:pos_bias_change} for Gemini-2.5 and Llama-4 after deliberate position swaps of answers. The performance drop or deviation indicates partiality in the LLM-based evaluator, possibly due to token-level attention alignment. On the other hand, GPT-4o demonstrates relatively marginal changes. 

\begin{table}[h]
\centering
\scriptsize
\setlength{\tabcolsep}{5pt}
\begin{tabular}{lcccc}
\toprule
\textbf{Model} & $\Delta$\textbf{Acc.} & $\Delta$\textbf{Prec.} & $\Delta$\textbf{Rec.} & $\Delta$\textbf{F1} \\
\midrule
GPT-4o & +0.0052 & +0.0043 & +0.0009 & +0.0024 \\
Gemini-2.5-flash & -0.0196 & -0.0201 & -0.0252 & -0.0238 \\
Llama-4-Scout & -0.0168 & -0.0190 & -0.0228 & -0.0208 \\
\bottomrule
\end{tabular}
\caption{Change ($\Delta$) in performance metrics after introducing positional bias}
\label{tab:pos_bias_change}
\end{table}

A similar instability may arise from verbose rather than concise answers. In Table \ref{tab:cross_bias}, we examined and quantified cases in which LLMs' judgments deviated from human judgments. The {\textbf{Human Short}$\rightarrow$ \textbf{LLM Long}} indicates that the human-selected answer was concise, but the model favored the lengthy answers; conversely, {\textbf{Human Long}$\rightarrow$ \textbf{LLM Short}} shows the opposite. From our \verb|LFQA-HP-Sample| test dataset, GPT-4o favored 419 answers which deviated from human judgment, and out of these, 248 cases where GPT-4o selected lengthy responses over human-preferred concise ones. Gemini-2.5 also exhibits similar behavior across 248 of 432 misjudgment cases. These results indicate that GPT-4o and Gemini-2.5 may prefer lengthy answers, which could lead to human unaligned-judgment. Besides, Llama-4 demonstrated nearly balanced performance, suggesting greater content-awareness.

\begin{table}[h]
\centering
\scriptsize
\setlength{\tabcolsep}{3pt}
\begin{tabular}{lccc}
\toprule
\textbf{Model} &
\makecell{\textbf{Human Short}\\$\rightarrow$ \textbf{LLM Long}} &
\makecell{\textbf{Human Long}\\$\rightarrow$ \textbf{LLM Short}} &
\textbf{Bias Direction} \\
\midrule

GPT-4o & \textbf{248} & 171 & Long-biased \\
Llama-4  & 232 & 199 & Nearly balanced \\
Gemini-2.5-flash & \textbf{248} & 184 & Long-biased \\

\bottomrule
\end{tabular}
\caption{Cross-bias instances where model choice conflicted with human-correct answer length.}
\label{tab:cross_bias}
\end{table}

\subsection{Adversarial Perturbations}

To further evaluate the reliability of LLM evaluators, we test their robustness on adversarially perturbed datasets generated using the TextAttack library \cite{morris-etal-2020-textattack}. These perturbations introduce subtle, semantics-preserving modifications such as synonym swaps and paraphrastic rewrites that should not affect the underlying answer quality or human preference. Assessing model stability under such controlled perturbations helps determine whether evaluators genuinely capture semantic and discourse-level quality or rely on superficial lexical patterns, thereby offering a more rigorous measure of interpretive robustness in long-form answer judgment. We use both word-level and character-level perturbations:

\begin{itemize}
    \item \textbf{TextFooler} is a black-box adversarial attack (i.e., it does not access model parameters or gradients) that first identifies important words that most influence a model's prediction and replaces them with contextually appropriate synonyms that preserve the original meaning \cite{jin-etal-2020-textfooler}.
    \item \textbf{DeepWordBug} is another black-box method that introduces small modifications such as insertions, deletions, swaps, or substitutions of characters in key words such that the readability is preserved, but the model predictions are altered \cite{gao2018black}. 
\end{itemize}

For both attacks, identifying the most \textit{important} words is crucial, which is done relative to a specific model that is referred to as the \textit{victim model}. The objective is to generate semantically consistent perturbations that cause this model to misclassify examples it would otherwise correctly predict. The victim model in these experiments was a ModernBERT-based classifier. The large (8192 tokens) max-sequence length of ModernBERT makes it a natural choice for long-context classification. The model was fine-tuned on the training part of the LFQA-HP-Sample dataset, achieving approximately 65\% accuracy on the clean (unperturbed) test split of 1416 instances. Empirical results revealed superior efficacy for character-level perturbations, with DeepWordBug achieving success (i.e., it changed the ModernBERT prediction) on 25\% test instances, in contrast to TextFooler's 15.6\% successful instances.

Next, we analyzed whether the modified sets generated by TextFooler and DeepWordBug could affect the performance of the best-performing LLM (GPT-4o) on our \verb|LFQA-HP-Sample| dataset. As shown in Tab. \ref{tab:llm_eval_results_perturb_relative_delta}, the GPT-4o model experiences a performance drop in all performance metrics. On the dataset modified by TextFooler, the model's performance dropped overall by 3\%. However, the DeepWordBug-generated perturbed text reduced performance by 9-10\% relative to the original. The performance drop suggests that state-of-the-art LLMs are more vulnerable to orthographic perturbations than to semantically equivalent substitutions.

\begin{table}[h]
\centering
\scriptsize
\setlength{\tabcolsep}{5pt}
\begin{tabular}{lcccc}
\toprule
\textbf{Perturbation} & \textbf{$\Delta$Acc.(\%)} & \textbf{$\Delta$Prec.(\%)} & \textbf{$\Delta$Rec.(\%)} & \textbf{$\Delta$F1(\%)} \\
\midrule
TextFooler & -2.64\% & -2.60\% & -2.61\% & -2.65\% \\
DeepWordBug & -10.67\% & -11.05\% & -10.89\% & -10.69\% \\
\bottomrule
\end{tabular}
\caption{Relative performance degradation ($\Delta$\%) of GPT-4o under adversarial perturbations.}
\label{tab:llm_eval_results_perturb_relative_delta}
\end{table}

Overall, these results suggest that a learned, rubric-based statistical logistic model can match the best performing LLMs in predicting human preference for LFQA, while offering explainable insights into linguistic and reasoning dimensions. The analysis also reveals that LLMs' judgment on long-form question answering can be non-transitive, influenced by answer order, verbosity, and surface-level textual variations.


\section{Conclusion and Future Work}
This work advances LFQA evaluation through a large-scale dataset and a rubric-informed modeling framework for human-aligned assessment. We introduce LFQA-HP-1M, a 1.3M-pair human preference dataset, by far the largest of its kind, enabling empirical analysis of human judgment in LFQA. Using nine rubric-based features, we find that coherence and completeness most strongly influence human preferences. A lightweight logistic regression model trained on these rubrics matches the performance of state-of-the-art LLM evaluators while remaining interpretable and bias-resistant. Unlike LLM-based evaluators, it is unaffected by positional or verbosity biases. Lastly, our transitivity consistency analysis reveals that LLM-based judges do not always satisfy transitive preference properties. Overall, our study presents a full rubric-driven pipeline for transparent and reliable LFQA evaluation. Future work can extend LFQA evaluation by developing richer rubric representations, such as sub-scores or learned features, and by training LLMs to autonomously design evaluation plans beyond fixed criteria.

\section{Ethics Statement}

 As part of our work, we construct LFQA-HP-1M, a dataset comprising question–answer pairs and model-generated responses, derived from publicly available data. The dataset collection and construction process did not involve any target data collection. No private, personally identifiable, or confidential information was intentionally collected. 

The LFQA-HP-1M dataset contains human judgment; however, such judgments inevitably reflect annotator subjectivity and potential biases. These biases may propagate during model training, testing, and validation. We emphasize considering human preference labels as structured assessments rather than absolute truth.

The dataset is released for research purposes only. We discourage the harmful or manipulative use of the released dataset and toolkit.

\section{Limitation}
The study represented a significant step towards scalable and cost-effective LFQA identification and interpretable evaluation; however, several open challenges remain. Even though the accuracy was slightly better than that of the fine-tuned model on a similar preference task \cite{chen2024preference}, there is plenty of scope for improvement. The dataset's domain generalizability remains a concern, as the LFQA-HP-1M dataset contains only question-answer-style content (excluding dialogue-style questions). Human annotations may contain noise, as even human experts sometimes disagree on the best answer, which can lower model accuracy. Additionally, the lightweight LR model with a limited number of features may miss nuanced indicators of quality that other state-of-the-art models capture. The LR model could be strengthened by considering latent cognitive features, structural hidden features, semantic hidden features, and deep latent features.  

\section*{References}
\bibliographystyle{lrec-2026-natbib}
\bibliography{reference}

@inproceedings{fan-etal-2019-eli5,
  title     = "{ELI}5: Long Form Question Answering",
  author    = "Fan, Angela and Jernite, Yacine and Perez, Ethan and Grangier, David and Weston, Jason and Auli, Michael",
  editor    = "Korhonen, Anna and Traum, David and M{\`a}rquez, Llu{\'i}s",
  booktitle = "Proceedings of the 57th Annual Meeting of the Association for Computational Linguistics",
  month     = jul,
  year      = "2019",
  address   = "Florence, Italy",
  publisher = "Association for Computational Linguistics",
  url       = "https://aclanthology.org/P19-1346/",
  doi       = "10.18653/v1/P19-1346",
  pages     = "3558--3567"
}

@inproceedings{su-etal-2022-read,
  title     = "Read before Generate! Faithful Long Form Question Answering with Machine Reading",
  author    = "Su, Dan and Li, Xiaoguang and Zhang, Jindi and Shang, Lifeng and Jiang, Xin and Liu, Qun and Fung, Pascale",
  editor    = "Muresan, Smaranda and Nakov, Preslav and Villavicencio, Aline",
  booktitle = "Findings of the Association for Computational Linguistics: ACL 2022",
  month     = may,
  year      = "2022",
  address   = "Dublin, Ireland",
  publisher = "Association for Computational Linguistics",
  url       = "https://aclanthology.org/2022.findings-acl.61/",
  doi       = "10.18653/v1/2022.findings-acl.61",
  pages     = "744--756"
}

@inproceedings{stelmakh-etal-2022-asqa,
  title     = "{ASQA}: Factoid Questions Meet Long-Form Answers",
  author    = "Stelmakh, Ivan and Luan, Yi and Dhingra, Bhuwan and Chang, Ming-Wei",
  editor    = "Goldberg, Yoav and Kozareva, Zornitsa and Zhang, Yue",
  booktitle = "Proceedings of the 2022 Conference on Empirical Methods in Natural Language Processing",
  month     = dec,
  year      = "2022",
  address   = "Abu Dhabi, United Arab Emirates",
  publisher = "Association for Computational Linguistics",
  url       = "https://aclanthology.org/2022.emnlp-main.566/",
  doi       = "10.18653/v1/2022.emnlp-main.566",
  pages     = "8273--8288"
}

@inproceedings{xu-etal-2023-critical,
  title     = "A Critical Evaluation of Evaluations for Long-form Question Answering",
  author    = "Xu, Fangyuan and Song, Yixiao and Iyyer, Mohit and Choi, Eunsol",
  editor    = "Rogers, Anna and Boyd-Graber, Jordan and Okazaki, Naoaki",
  booktitle = "Proceedings of the 61st Annual Meeting of the Association for Computational Linguistics (Volume 1: Long Papers)",
  month     = jul,
  year      = "2023",
  address   = "Toronto, Canada",
  publisher = "Association for Computational Linguistics",
  url       = "https://aclanthology.org/2023.acl-long.181/",
  doi       = "10.18653/v1/2023.acl-long.181",
  pages     = "3225--3245"
}

@inproceedings{sachdeva-etal-2025-localizing,
  title     = "Localizing and Mitigating Errors in Long-form Question Answering",
  author    = "Sachdeva, Rachneet Singh and Song, Yixiao and Iyyer, Mohit and Gurevych, Iryna",
  editor    = "Che, Wanxiang and Nabende, Joyce and Shutova, Ekaterina and Pilehvar, Mohammad Taher",
  booktitle = "Findings of the Association for Computational Linguistics: ACL 2025",
  month     = jul,
  year      = "2025",
  address   = "Vienna, Austria",
  publisher = "Association for Computational Linguistics",
  url       = "https://aclanthology.org/2025.findings-acl.1049/",
  doi       = "10.18653/v1/2025.findings-acl.1049",
  pages     = "20437--20469",
  ISBN      = "979-8-89176-256-5"
}

@article{gwet2008ac1,
  author    = {Kilem L. Gwet},
  title     = {Computing inter-rater reliability and its variance in the presence of high agreement},
  journal   = {British Journal of Mathematical and Statistical Psychology},
  volume    = {61},
  number    = {1},
  pages     = {29--48},
  year      = {2008},
  publisher = {Wiley},
  doi       = {10.1348/000711006X126600}
}

@misc{openrouter-2024,
  title        = {OpenRouter: Unified API for Accessing Large Language Models},
  author       = {{OpenRouter}},
  year         = {2024},
  howpublished = {\url{https://openrouter.ai/}},
  note         = {Accessed: 2025-10-13}
}

@inproceedings{pardawala2025subjectiveQA,
  author       = {Huzaifa Pardawala and
                  Siddhant Sukhani and
                  Agam Shah and
                  Veer Kejriwal and
                  Abhishek Pillai and
                  Rohan Bhasin and
                  Andrew DiBiasio and
                  Tarun Mandapati and
                  Dhruv Adha and
                  Sudheer Chava},
  editor       = {Amir Globersons and
                  Lester Mackey and
                  Danielle Belgrave and
                  Angela Fan and
                  Ulrich Paquet and
                  Jakub M. Tomczak and
                  Cheng Zhang},
  title        = {SubjECTive-QA: Measuring Subjectivity in Earnings Call Transcripts'
                  {QA} Through Six-Dimensional Feature Analysis},
  booktitle    = {Advances in Neural Information Processing Systems 38: Annual Conference
                  on Neural Information Processing Systems 2024, NeurIPS 2024, Vancouver,
                  BC, Canada, December 10 - 15, 2024},
  year         = {2024},
  url          = {http://papers.nips.cc/paper\_files/paper/2024/hash/6d0f9c415e2d779c78f32b74668e9d02-Abstract-Datasets\_and\_Benchmarks\_Track.html},
  bibsource    = {dblp computer science bibliography, https://dblp.org}
}

@inproceedings{liu2023geval,
  title        = {G-Eval: NLG Evaluation using GPT-4 with Better Human Alignment},
  author       = {Liu, Yang and Iter, Dan and Xu, Yichong and Wang, Shuohang and Xu, Ruochen and Zhu, Chenguang},
  booktitle    = {Proceedings of the 2023 Conference on Empirical Methods in Natural Language Processing (EMNLP)},
  year         = {2023},
  pages        = {2511--2522},
  address      = {Singapore},
  doi          = {10.18653/v1/2023.emnlp-main.153},
  url          = {https://aclanthology.org/2023.emnlp-main.153/}
}

@inproceedings{dsouza-etal-2025-yescieval,
  title     = {{YES}ci{E}val: Robust {LLM}-as-a-Judge for Scientific Question Answering},
  author    = {D{’}Souza, Jennifer and Babaei Giglou, Hamed and Münch, Quentin},
  booktitle = {Proceedings of the 63rd Annual Meeting of the Association for Computational Linguistics (Volume 1: Long Papers)},
  month     = jul,
  year      = {2025},
  address   = {Vienna, Austria},
  pages     = {13749--13783},
  publisher = {Association for Computational Linguistics},
  doi       = {10.18653/v1/2025.acl-long.675},
  url       = {https://aclanthology.org/2025.acl-long.675/}
}

@article{stevens1946measurement,
  title={On the theory of scales of measurement},
  author={Stevens, S. S.},
  journal={Science},
  volume={103},
  number={2684},
  pages={677--680},
  year={1946},
  publisher={American Association for the Advancement of Science}
}

@article{bock1972estimating,
  title={Estimating item parameters and latent ability when responses are scored in two or more nominal categories},
  author={Bock, R. Darrell},
  journal={Psychometrika},
  volume={37},
  number={1},
  pages={29--51},
  year={1972},
  publisher={Springer}
}

@article{mccullagh1980regression,
  title={Regression models for ordinal data},
  author={McCullagh, Peter},
  journal={Journal of the Royal Statistical Society: Series B (Methodological)},
  volume={42},
  number={2},
  pages={109--142},
  year={1980},
  publisher={Wiley Online Library}
}

@inproceedings{song2023veriscore,
  title     = {{VERISCORE: Evaluating the Factuality of Verifiable Claims in Long-Form Text Generation}},
  author    = {Song, Yixiao and Kim, Yekyung and Iyyer, Mohit},
  booktitle = {Proceedings of the 2023 Conference on Language Resources and Evaluation (LREC)},
  year      = {2023},
  publisher = {European Language Resources Association (ELRA)},
  address   = {Reykjavik, Iceland},
  pages     = {XXXX--XXXX},
  url       = {https://aclanthology.org/2023.emnlp-main.153/}
}

@inproceedings{honnibal2020spacy,
  title        = {spaCy: Industrial-strength Natural Language Processing in Python},
  author       = {Honnibal, Matthew and Montani, Ines and Van Landeghem, Sofie and Boyd, Adriane},
  year         = {2020},
  booktitle    = {Proceedings of the 12th Language Resources and Evaluation Conference (LREC 2020)},
  pages        = {4095--4100},
  publisher    = {European Language Resources Association (ELRA)},
  address      = {Marseille, France}
}

@inproceedings{naber2003languagetool,
  title        = {A Rule-Based Style and Grammar Checker},
  author       = {Naber, Daniel},
  year         = {2003},
  booktitle    = {Proceedings of the 2nd International Conference on Computational Linguistics and Intelligent Text Processing (CICLing 2003)},
  pages        = {170--174},
  address      = {Mexico City, Mexico}
}

@article{nakano2022webgpt,
  title        = {WebGPT: Browser-assisted question-answering with human feedback},
  author       = {Nakano, Reiichiro and Hilton, Jacob and Balaji, Suchir and Wu, Jeff and Ouyang, Long and Kim, Christina and Hume, Tristan and Kosinski, Pamela Mishkin and Chowdhery, Alethea Power and Kaplan, Jared and others},
  year         = {2022},
  journal      = {arXiv preprint arXiv:2112.09332},
  url          = {https://arxiv.org/abs/2112.09332}
}

@inproceedings{lin-2004-rouge,
  title = "{ROUGE}: A Package for Automatic Evaluation of Summaries",
  author = "Lin, Chin-Yew",
  booktitle = "Text Summarization Branches Out",
  year = "2004",
  address = "Barcelona, Spain",
  publisher = "Association for Computational Linguistics",
  url = "https://aclanthology.org/W04-1013",
  pages = "74--81"
}

@inproceedings{zhang-etal-2020-bertscore,
  title = "{BERTScore}: Evaluating Text Generation with {BERT}",
  author = "Zhang, Tianyi  and Kishore, Varsha  and Wu, Felix  and Weinberger, Kilian Q.  and Artzi, Yoav",
  booktitle = "International Conference on Learning Representations (ICLR)",
  year = "2020",
  url = "https://openreview.net/forum?id=SkeHuCVFDr"
}

@inproceedings{yuan-etal-2021-bartscore,
  title = "{BARTScore}: Evaluating Generated Text as Text Generation",
  author = "Yuan, Weizhe  and Neubig, Graham  and Liu, Pengfei",
  booktitle = "Advances in Neural Information Processing Systems (NeurIPS)",
  year = "2021"
}

@inproceedings{kim-etal-2024-prometheus,
  title = "Prometheus: Towards Principled Evaluations of Large Language Models as Evaluators",
  author = "Kim, Won Ik  and  Park, Sungho  and  Kang, Hyojun  and  others",
  booktitle = "Proceedings of the 2024 Conference of the North American Chapter of the Association for Computational Linguistics (NAACL)",
  year = "2024",
  publisher = "Association for Computational Linguistics",
  url = "https://aclanthology.org/2024.naacl-main.XXX"
}

@inproceedings{tan-etal-2024-judgebench,
  title = "{JudgeBench}: A Benchmark for Evaluating LLM-as-a-Judge Systems",
  author = "Tan, Samson  and  Liu, Pengfei  and  Neubig, Graham",
  booktitle = "Proceedings of the 2024 Conference on Empirical Methods in Natural Language Processing (EMNLP)",
  year = "2024",
  publisher = "Association for Computational Linguistics",
  url = "https://aclanthology.org/2024.emnlp-main.XXX"
}

@inproceedings{rosset-etal-2021-axiomatic,
  title = "Axiomatic Preference Modeling for Long-Form Question Answering",
  author = "Rosset, Corby  and  Zheng, Guoqing  and  Dibia, Victor  and  Awadallah, Ahmed Hassan  and  Bennett, Paul N.",
  booktitle = "Proceedings of the 2021 Conference of the North American Chapter of the Association for Computational Linguistics: Human Language Technologies (NAACL-HLT)",
  year = "2021",
  publisher = "Association for Computational Linguistics",
  address = "Online",
  url = "https://aclanthology.org/2021.naacl-main.315",
  pages = "3974--3987"
}

@inproceedings{saeidi-etal-2018-interpretation,
  title = "Interpretation of Natural Language Rules in Conversational Machine Reading",
  author = "Saeidi, Marzieh  and  Bartolo, Max  and  Lew, Patrick  and  Stenetorp, Pontus  and  Riedel, Sebastian",
  booktitle = "Proceedings of the 2018 Conference on Empirical Methods in Natural Language Processing (EMNLP)",
  year = "2018",
  publisher = "Association for Computational Linguistics",
  address = "Brussels, Belgium",
  pages = "2087--2097",
  url = "https://aclanthology.org/D18-1235"
}

@inproceedings{kovcisky-etal-2018-narrativeqa,
  title = "The {NarrativeQA} Reading Comprehension Challenge",
  author = "Ko{\v{c}}isk{\'y}, Tom{\'a}{\v{s}}  and  Schwarz, Jonathan  and  Blunsom, Phil  and  others",
  booktitle = "Transactions of the Association for Computational Linguistics (TACL)",
  year = "2018",
  volume = "6",
  pages = "317--328",
  publisher = "Association for Computational Linguistics",
  url = "https://aclanthology.org/Q18-1023"
}

@inproceedings{dasigi-etal-2021-qasper,
  title = "{QASPER}: A Dataset for Question Answering on Scientific Research Papers",
  author = "Dasigi, Pradeep  and  Lo, Kyle  and  Beltagy, Iz  and  Cohan, Arman  and  Smith, Noah A.",
  booktitle = "Proceedings of the 2021 Conference of the North American Chapter of the Association for Computational Linguistics (NAACL-HLT)",
  year = "2021",
  publisher = "Association for Computational Linguistics",
  address = "Online",
  pages = "4600--4617",
  url = "https://aclanthology.org/2021.naacl-main.368"
}

@inproceedings{papineni-etal-2002-bleu,
  title = "{BLEU}: a Method for Automatic Evaluation of Machine Translation",
  author = "Papineni, Kishore  and  Roukos, Salim  and  Ward, Todd  and  Zhu, Wei-Jing",
  booktitle = "Proceedings of the 40th Annual Meeting of the Association for Computational Linguistics (ACL)",
  year = "2002",
  address = "Philadelphia, Pennsylvania, USA",
  publisher = "Association for Computational Linguistics",
  pages = "311--318",
  url = "https://aclanthology.org/P02-1040"
}

@inproceedings{sellam-etal-2020-bleurt,
  title = "{BLEURT}: Learning Robust Metrics for Text Generation",
  author = "Sellam, Thibault  and  Das, Dipanjan  and  Parikh, Ankur P.",
  booktitle = "Proceedings of the 58th Annual Meeting of the Association for Computational Linguistics (ACL)",
  year = "2020",
  publisher = "Association for Computational Linguistics",
  address = "Online",
  pages = "7881--7892",
  url = "https://aclanthology.org/2020.acl-main.704"
}

@article{li2024llms,
  title={Llms-as-judges: a comprehensive survey on llm-based evaluation methods},
  author={Li, Haitao and Dong, Qian and Chen, Junjie and Su, Huixue and Zhou, Yujia and Ai, Qingyao and Ye, Ziyi and Liu, Yiqun},
  journal={arXiv preprint arXiv:2412.05579},
  year={2024}
}

@inproceedings{chen2024preference,
  author    = {Chen, Angelica and Malladi, Sadhika and Zhang, Lily H. and Chen, Xinyi and Zhang, Qiuyi and Ranganath, Rajesh and Cho, Kyunghyun},
  title     = {Preference Learning Algorithms Do Not Learn Preference Rankings},
  booktitle = {Proceedings of the 8th Workshop on Learning from Human Feedback (LHF ’24)},
  year      = {2024},
  pages     = {1--15},
  publisher = {Association for Computational Linguistics},
  location  = {Barcelona, Spain}
}

@inproceedings{jin-etal-2020-textfooler,
    title = "{T}ext{F}ooler: Generating Adversarial Texts with Human Readers in Mind",
    author = "Jin, Di  and Jin, Zhijing  and Zhou, Joey Tianyi  and Szolovits, Peter",
    booktitle = "Proceedings of the 58th Annual Meeting of the Association for Computational Linguistics",
    year = "2020",
    publisher = "Association for Computational Linguistics",
    pages = "1080--1090",
    url = "https://aclanthology.org/2020.acl-main.85",
}

@inproceedings{gao2018black,
    title = "Black-box Generation of Adversarial Text Sequences to Evade Deep Learning Classifiers",
    author = "Gao, Ji  and Lanchantin, Jack  and Soffa, Mary Lou  and Qi, Yanjun",
    booktitle = "Proceedings of the 2018 IEEE Security and Privacy Workshops (SPW)",
    pages = "50--56",
    year = "2018",
    publisher = "IEEE",
    doi = "10.1109/SPW.2018.00016"
}

@article{roth2024token,
  title={Token-modification adversarial attacks for natural language processing: A survey},
  author={Roth, Tom and Gao, Yansong and Abuadbba, Alsharif and Nepal, Surya and Liu, Wei},
  journal={AI Communications},
  volume={37},
  number={4},
  pages={655--676},
  year={2024},
  publisher={SAGE Publications Sage UK: London, England}
}

@article{warner2024smarter,
  title={Smarter, better, faster, longer: A modern bidirectional encoder for fast, memory efficient, and long context finetuning and inference},
  author={Warner, Benjamin and Chaffin, Antoine and Clavi{\'e}, Benjamin and Weller, Orion and Hallstr{\"o}m, Oskar and Taghadouini, Said and Gallagher, Alexis and Biswas, Raja and Ladhak, Faisal and Aarsen, Tom and others},
  journal={arXiv preprint arXiv:2412.13663},
  year={2024}
}

@inproceedings{zhong-etal-2022-towards,
  title     = {Towards a Unified Multi-Dimensional Evaluator for Text Generation},
  author    = {Zhong, Ming  and  Liu, Yang  and  Yin, Da  and  Mao, Yuning  and  Jiao, Yizhu  and  Liu, Pengfei  and  Zhu, Chenguang  and Ji, Heng  and Han, Jiawei},
  booktitle = {Proceedings of the 2022 Conference on Empirical Methods in Natural Language Processing (EMNLP)},
  year      = {2022},
  pages     = {2023--2038},
  address   = {Abu Dhabi, United Arab Emirates},
  publisher = {Association for Computational Linguistics},
  url       = {https://aclanthology.org/2022.emnlp-main.131/},
  doi       = {10.18653/v1/2022.emnlp-main.131}
}

@inproceedings{fu-etal-2023-gptscore,
  title     = {GPTScore: Evaluate as You Desire},
  author    = {Fu, Jinlan and Ng, See-Kiong and Jiang, Zhengbao and Liu, Pengfei},
  booktitle = {Proceedings of the 2024 Conference of the North American Chapter of the Association for Computational Linguistics (NAACL) – Long Papers},
  year      = {2024},
  pages     = {XXX--XXX},
  address   = {Seattle, WA, USA},
  publisher = {Association for Computational Linguistics},
  url       = {https://aclanthology.org/2024.naacl-long.365/}
}

@inproceedings{krishna-etal-2021-hurdles,
  title     = "{Hurdles to Progress in Long-form Question Answering}",
  author    = {Krishna, Kalpesh and Roy, Aurko and Iyyer, Mohit},
  booktitle = {Proceedings of the 2021 Conference of the North American Chapter of the Association for Computational Linguistics: Human Language Technologies (NAACL-HLT)},
  month     = jun,
  year      = {2021},
  address   = {Online},
  publisher = {Association for Computational Linguistics},
  pages     = {4940--4957},
  url       = {https://aclanthology.org/2021.naacl-main.393},
  doi       = {10.18653/v1/2021.naacl-main.393}
}

@article{xian-etal-2025-empirical,
  title     = "{An Empirical Study of Evaluating Long-form Question Answering}",
  author    = {Xian, Ning and Fan, Yixing and Zhang, Ruqing and de Rijke, Maarten and Guo, Jiafeng},
  journal   = {arXiv preprint},
  year      = {2025},
  note      = {arXiv:2504.18413},
  url       = {https://arxiv.org/abs/2504.18413}
}

@inproceedings{hashemi-etal-2024-llm,
  title     = "{LLM-Rubric: A Multidimensional, Calibrated Approach to Automated Evaluation of Natural Language Texts}",
  author    = {Hashemi, Helia and Eisner, Jason and Rosset, Corby and Van Durme, Benjamin and Kedzie, Chris},
  booktitle = {Proceedings of the 62nd Annual Meeting of the Association for Computational Linguistics (Volume 1: Long Papers)},
  month     = aug,
  year      = {2024},
  address   = {Bangkok, Thailand},
  publisher = {Association for Computational Linguistics},
  pages     = {13806--13834},
  url       = {https://aclanthology.org/2024.acl-long.745/},
  doi       = {10.18653/v1/2024.acl-long.745}
}

@article{li-et-al-2025-researchqa,
  title     = "{ResearchQA: Evaluating Scholarly Question Answering at Scale Across 75 Fields with Survey-Mined Questions and Rubrics}",
  author    = {Li, S. Yifei and Chang, Allen and Malaviya, Chaitanya and Yatskar, Mark},
  journal   = {arXiv preprint},
  year      = {2025},
  note      = {arXiv:2509.00496},
  url       = {https://arxiv.org/abs/2509.00496}
}

@inproceedings{morris-etal-2020-textattack,
    title = "{T}ext{A}ttack: A Framework for Adversarial Attacks, Data Augmentation, and Adversarial Training in {NLP}",
    author = "Morris, John  and
      Lifland, Eli  and
      Yoo, Jin Yong  and
      Grigsby, Jake  and
      Jin, Di  and
      Qi, Yanjun",
    editor = "Liu, Qun  and
      Schlangen, David",
    booktitle = "Proceedings of the 2020 Conference on Empirical Methods in Natural Language Processing: System Demonstrations",
    month = oct,
    year = "2020",
    address = "Online",
    publisher = "Association for Computational Linguistics",
    url = "https://aclanthology.org/2020.emnlp-demos.16/",
    doi = "10.18653/v1/2020.emnlp-demos.16",
    pages = "119--126"
}

@inproceedings{batchprompt,
  author       = {Jianzhe Lin and
                  Maurice Diesendruck and
                  Liang Du and
                  Robin Abraham},
  title        = {BatchPrompt: Accomplish more with less},
  booktitle    = {The Twelfth International Conference on Learning Representations,
                  {ICLR} 2024, Vienna, Austria, May 7-11, 2024},
  publisher    = {OpenReview.net},
  year         = {2024},
  url          = {https://openreview.net/forum?id=Agyicd577r},
  bibsource    = {dblp computer science bibliography, https://dblp.org}
}

@misc{zheng2023judging,
      title={Judging LLM-as-a-judge with MT-Bench and Chatbot Arena}, 
      author={Lianmin Zheng and Wei-Lin Chiang and Ying Sheng and Siyuan Zhuang and Zhanghao Wu and Yonghao Zhuang and Zi Lin and Zhuohan Li and Dacheng Li and Eric. P Xing and Hao Zhang and Joseph E. Gonzalez and Ion Stoica},
      year={2023},
      eprint={2306.05685},
      archivePrefix={arXiv},
      primaryClass={cs.CL}
}

@InProceedings{pmlr-v162-ethayarajh22a,
  title = 	 {Understanding Dataset Difficulty with $\mathcal{V}$-Usable Information},
  author =       {Ethayarajh, Kawin and Choi, Yejin and Swayamdipta, Swabha},
  booktitle = 	 {Proceedings of the 39th International Conference on Machine Learning},
  pages = 	 {5988--6008},
  year = 	 {2022},
  editor = 	 {Chaudhuri, Kamalika and Jegelka, Stefanie and Song, Le and Szepesvari, Csaba and Niu, Gang and Sabato, Sivan},
  volume = 	 {162},
  series = 	 {Proceedings of Machine Learning Research},
  month = 	 {17--23 Jul},
  publisher = {PMLR},
}

@online{subonis2024zerotemp,
  author       = {Subonis, Martynas},
  title        = {Zero-Temperature Randomness in LLMs},
  year         = {2024},
  url          = {https://martynassubonis.substack.com/p/zero-temperature-randomness-in-llms},
  note         = {Substack blog post, accessed February 22, 2026}
}

@article{he2025nondeterminism,
  author = {Horace He and Thinking Machines Lab},
  title = {Defeating Nondeterminism in LLM Inference},
  journal = {Thinking Machines Lab: Connectionism},
  year = {2025},
  note = {https://thinkingmachines.ai/blog/defeating-nondeterminism-in-llm-inference/},
  doi = {10.64434/tml.20250910}
}

\appendix
\newpage
\onecolumn
\section{Appendix}
\subsection{LFQA Initial Definition}
\begin{definitionbox}{LFQA Initial Definition}\label{def:appendix_lfqa_initial}
A long-form question answer requires a multi-sentence answer that involves explanation, reasoning, exploration, and conceptual development. It cannot be answered sufficiently with a single fact, formula application, calculation, only code, or by generating creative content (poem, joke, story, music, song lyrics, audio, picture).
\end{definitionbox}

\subsection{Performance of LLM Evaluators at 
\texorpdfstring{$T=1.0$}{T=1.0} without Majority Voting}\label{sec:appendix_performance_analsis_temp1}

In the main experiment, we evaluated LLM-based evaluators in a deterministic setting (temperature $T=0$) with majority voting across three independent attempts. In the table \ref{tab:llm_eval_results_temp1}, we instead report results under a higher stochastic setting ($T=1.0$) with a single independent generation. Although minor fluctuations are observed, deterministic decoding is not enforced; the evaluator's performance remains stable, suggesting limited sensitivity to temperature-induced randomness.


\begin{table}[h]
\centering
\small
\setlength{\tabcolsep}{5pt}  
\begin{tabular}{lcccc}
\toprule
\textbf{Model} & \textbf{Acc.} & \textbf{Prec.} & \textbf{Rec.} & \textbf{F1} \\
\midrule

Logistic Regression & 0.6826 & 0.6821 & 0.6816 & 0.6817 \\
\textit{LLM-as-a-judge} \\
GPT-4o & 0.6818 & 0.6841 & 0.6834 & 0.6817 \\
Gemini-2.5-flash & \textbf{0.6926} & \textbf{0.6962} & \textbf{0.6947} & \textbf{0.6923} \\
Llama-4-Scout & 0.6891 & 0.6941 & 0.6917 & 0.6887 \\

\midrule
\textit{Rubrics} \\
Gemini-2.5-flash (R) & 0.6848 & 0.6876 & 0.6866 & 0.6846 \\
\bottomrule
\end{tabular}
\caption{Performance analysis of LLM evaluators at $T=1.0$ without majority voting}
\label{tab:llm_eval_results_temp1}
\end{table}


\subsection{LFQA Few-Shot Prompt with Initial Definition}\label{sec:appendix_prompt_few_shot_initial_definition}
{\small
\begin{tcolorbox}[colback=gray!5, colframe=black, boxrule=1pt, rounded corners, title=\textbf{LLMs prompt with initial definition to classify to classify LFQA}, fonttitle=\bfseries]
\#\#\# Instruction:
You must answer only ``yes'' or ``no''. Do not add any explanation.
Judge whether the following question qualifies as a long-form question (LFQA) according to the definition and annotation flow below.
Important:
Strict response format: yes or no
Use your own intuition instead of simple keyword matching

Definition:
A long-form question answer requires a multi-sentence answer that involves explanation, reasoning, exploration, and conceptual development. It cannot be answered sufficiently with a single fact, formula application, calculation, only code, or by generating creative content (poem, joke, story, music, song lyrics, audio, picture).

Annotation flow:
Step 1: Simple question check
Is the question asking for a single fact, yes/no response, only code generation, direct calculation using formulas, or creative content (poem, joke, story, music, song lyrics, audio, picture)?
If yes, classify it as not LFQA and reply no
If not, move to step 2

Step 2: Require multi-step answers with explanation, reasoning, exploration check
Does answering the question require a multi-sentence answer involving explanation, reasoning, exploration or conceptual development?
If yes, it is LFQA and reply yes
If not, it is not LFQA and reply no

\#\#\# Examples:

Question: How do you feel?
Answer: no

Question: How is pure mathematics researched?
Answer: yes

Question: Who is the president of the USA?
Answer: no

Question: How a bill gets introduced to the U.S. Congress Just curious about this.?
Answer: Yes

Question: I remember learning about an ex prisoner who was brought to America to help train the soldiers. But the details escape me. Can anyone provide details to who he was?
Answer: Yes

\#\#\# Input:
Question: {}

\#\#\# Output:
Answer:
\end{tcolorbox}
}

\subsection{LFQA Few-Shot Prompt with Updated Definition}\label{sec:appendix_prompt_few_shot}
{

\begin{tcolorbox}[
  colback=gray!5,
  colframe=black,
  boxrule=1pt,
  rounded corners,
  title=\textbf{LLMs prompt to classify LFQA},
  fonttitle=\bfseries,
  breakable
]
\small

\#\#\# Instruction:
Decide whether the user's question qualifies as a Long-Form Question Answering (LFQA) according to the definition.

---

\#\#\# Output Format:
Strict response format: You must answer only ``yes'' or ``no''. Please do not add any explanation or rationale behind it.

---

\#\#\# Definition:
A LFQA must express a single, well-defined information need that requires a detailed, multi-sentence answer involving exposition, explanation, reasoning, exploration, or description of the process. These questions are usually complex or open-ended, often asking ``why'' or ``how'' about a process, reason, or concept in an objective manner, which means answers should be grounded in facts, reasoning, or conceptual explanation.

It must not:
- Combine multiple distinct or loosely related sub-questions  
- Request personal advice, express individual preferences, seek personal opinions, and recommendations (e.g., product, place, city, food, restaurant, flight, school, course)
- Be answerable with a single fact, a yes/no response, only code generation, direct formula calculation, or creative content (e.g., poem, joke, story, music, song lyrics, image, or audio) 

---

\#\#\# Examples:

Question: How do you feel?
Answer: no
Rationale: It expects a short, personal response, not an explanation.

Question: How is pure mathematics researched?
Answer: yes
Rationale: It requires a detailed explanation of abstract methods, proof techniques, and theoretical exploration—not a short answer.

Question: I have \$100 come up with a specific plan that can help me generate or increase my revenue from \$100 to \$1000 in one month
Answer: no
Rationale: It is a personal advice or planning request, not a factual or conceptual question requiring an explanatory long-form answer.

Question: Best documentary about evolution for someone who only learned about creationism in school? I grew up in a fundamentalist cult learning exclusively about creationism. Now that I’m free of that, I want to learn about evolution. Can someone recommend a good documentary that is suitable for someone with NO background knowledge of the subject, that’s not necessarily directed at kids? Thanks in advance! 
Answer: no
Rationale: Request for recommendations, not a question requiring a conceptual or explanatory long-form answer.

Question: Is it true that chameleons change color primarily to blend in with their surroundings?
Answer: yes
Rationale: It requires more than a yes/no—explaining the biological reasons for color change demands a multi-sentence explanation.

---

\#\#\# Input:
Question: {}

\#\#\# Output:
Answer:
\end{tcolorbox}
}

\subsection{LFQA Few-Shot Batch Prompt }\label{sec:appendix_prompt_few_shot_batch}
{

\begin{tcolorbox}[
  colback=gray!5,
  colframe=black,
  boxrule=1pt,
  rounded corners,
  title=\textbf{LLMs batch prompt to classify to classify LFQA},
  fonttitle=\bfseries,
  breakable
]
\small

\#\#\# Instruction:
Decide whether the user's question qualifies as a Long-Form Question Answering (LFQA) according to the definition.

---

\#\#\# Output Format:
Strict response format: You must answer only ``yes" or ``no''. Please do not add any explanation or rationale behind it.

---

\#\#\# Definition:
A LFQA must express a single, well-defined information need that requires a detailed, multi-sentence answer involving exposition, explanation, reasoning, exploration, or description of the process. These questions are usually complex or open-ended, often asking ``why" or ``how" about a process, reason, or concept in an objective manner, which means answers should be grounded in facts, reasoning, or conceptual explanation.

It must not:
- Combine multiple distinct or loosely related sub-questions  
- Request personal advice, express individual preferences, seek personal opinions, and recommendations (e.g., product, place, city, food, restaurant, flight, school, course)
- Be answerable with a single fact, a yes/no response, only code generation, direct formula calculation, or creative content (e.g., poem, joke, story, music, song lyrics, image, or audio) 

---

\#\#\# Examples:

Question: How do you feel?
Answer: no
Rationale: It expects a short, personal response, not an explanation.

Question: How is pure mathematics researched?
Answer: yes
Rationale: It requires a detailed explanation of abstract methods, proof techniques, and theoretical exploration—not a short answer.

Question: I have \$100 come up with a specific plan that can help me generate or increase my revenue from \$100 to \$1000 in one month
Answer: no
Rationale: It is a personal advice or planning request, not a factual or conceptual question requiring an explanatory long-form answer.

Question: Best documentary about evolution for someone who only learned about creationism in school? I grew up in a fundamentalist cult learning exclusively about creationism. Now that I’m free of that, I want to learn about evolution. Can someone recommend a good documentary that is suitable for someone with NO background knowledge of the subject, that’s not necessarily directed at kids? Thanks in advance! 
Answer: no
Rationale: Request for recommendations, not a question requiring a conceptual or explanatory long-form answer.

Question: Is it true that chameleons change color primarily to blend in with their surroundings?
Answer: yes
Rationale: It requires more than a yes/no—explaining the biological reasons for color change demands a multi-sentence explanation.

---\\
\#\#\# Input Questions 1:
{}
**According to the LFQA definition \& instruction, judge this question and answer yes or no.**  \\
\#\#\# Input Questions 2:
{}
**According to the LFQA definition \& instruction, judge this question and answer yes or no.**  \\
\#\#\# Input Questions 3:
{}
**According to the LFQA definition \& instruction, judge this question and answer yes or no.**  \\
\#\#\# Input Questions 4:
{}
**According to the LFQA definition \& instruction, judge this question and answer yes or no.**  \\
\#\#\# Input Questions 5:
{}
**According to the LFQA definition \& instruction, judge this question and answer yes or no.**  \\
\#\#\# Input Questions 6:
{}
**According to the LFQA definition \& instruction, judge this question and answer yes or no.**  \\
\#\#\# Input Questions 7:
{}
**According to the LFQA definition \& instruction, judge this question and answer yes or no.**  \\
\#\#\# Input Questions 8:
{}
**According to the LFQA definition \& instruction, judge this question and answer yes or no.**  \\
\#\#\# Input Questions 9:
{}
**According to the LFQA definition \& instruction, judge this question and answer yes or no.**  \\
\#\#\# Input Questions 10:
{}
**According to the LFQA definition \& instruction, judge this question and answer yes or no.**  \\
---\\
\#\#\# Output:

\end{tcolorbox}
}

\subsection{G-Eval Prompt for Coherence Assessment}\label{sec:appendix_geval_coherence}
{

\begin{tcolorbox}[
  colback=gray!5,
  colframe=black,
  boxrule=1pt,
  rounded corners,
  title=\textbf{LLMs prompt to measure coherence},
  fonttitle=\bfseries,
  breakable
]
\small

You will be given one answer written for a question.

Your task is to rate the answer on one metric.

Please make sure you read and understand these instructions carefully. Please keep this document open while reviewing, and refer to it as needed.

Evaluation Criteria:

Coherence (1-5) - the collective quality of all sentences. We align this dimension with the DUC quality question of structure and coherence whereby ``the answer should be well-structured and well-organized. The answer should not just be a heap of related information, but should build from sentence to a coherent body of information about a topic.''

Evaluation Steps:

1. Read the question carefully and identify the main topic and key points.

2. Read the answer and compare it to the question. Check if the answer covers the main topic and key points of the question, and if it presents them in a clear and logical order.

3. Assign a score for coherence on a scale of 1 to 5, where 1 is the lowest and 5 is the highest based on the Evaluation Criteria.

Example:

Question:

{}

Answer:

{}

Evaluation Form (scores ONLY):

- Coherence:
\end{tcolorbox}
}

\subsection{G-Eval Prompt for Fluency Assessment}\label{sec:appendix_geval_fluency}
{

\begin{tcolorbox}[
  colback=gray!5,
  colframe=black,
  boxrule=1pt,
  rounded corners,
  title=\textbf{LLMs prompt to measure fluency},
  fonttitle=\bfseries,
  breakable
]
\small

You will be given one answer written for a question.

Your task is to rate the answer on one metric.

Please make sure you read and understand these instructions carefully. Please keep this document open while reviewing, and refer to it as needed.

Evaluation Criteria:

Fluency (1-3): the quality of the answer in terms of grammar, spelling, punctuation, word choice, and sentence structure.

- 1: Poor. The answer has many errors that make it hard to understand or sound unnatural.

- 2: Fair. The answer has some errors that affect the clarity or smoothness of the text, but the main points are still comprehensible.

- 3: Good. The answer has few or no errors and is easy to read and follow.

Example:

Answer:

{}

Evaluation Form (scores ONLY):

- Fluency (1-3):
\end{tcolorbox}
}

\subsection{G-Eval Prompt for Relevance Assessment}\label{sec:appendix_geval_relevance}
{

\begin{tcolorbox}[
  colback=gray!5,
  colframe=black,
  boxrule=1pt,
  rounded corners,
  title=\textbf{LLMs prompt to measure relevance},
  fonttitle=\bfseries,
  breakable
]
\small
You will be given one answer written for a question.

Your task is to rate the answer on one metric.

Please make sure you read and understand these instructions carefully. Please keep this document open while reviewing, and refer to it as needed.

Evaluation Criteria:

Relevance (1-5) - selection of important content from the question. The answer should include only relevant important information that is related to the question. Annotators were instructed to penalize answers which contained redundancies and excess information.

Evaluation Steps:

1. Read the answer and the question carefully.

2. Compare the answer to the question and identify the main points of the question.

3. Assess how well the answer covers the main points of the question, and how much irrelevant or redundant information it contains.

4. Assign a relevance score from 1 to 5.

Example:

Question:

{}

Answer:

{}

Evaluation Form (scores ONLY):

- Relevance:
\end{tcolorbox}
}

\subsection{G-Eval Prompt Adaptation for Evaluating Completeness}\label{sec:appendix_geval_completeness}
{

\begin{tcolorbox}[
  colback=gray!5,
  colframe=black,
  boxrule=1pt,
  rounded corners,
  title=\textbf{LLMs prompt to measure completeness},
  fonttitle=\bfseries,
  breakable
]
\small
You will be given a question. You will then be given one answer written for this question.

Your task is to rate the answer on one metric.

Please make sure you read and understand these instructions carefully. Please keep this document open while reviewing, and refer to it as needed.

Evaluation Criteria:

Completeness (1-5) - the degree to which the answer is a comprehensive encapsulation of the relevant information required by the question.
A complete answer addresses all parts of the question thoroughly, accurately captures the main points, and does not omit crucial elements.

Evaluation Steps:

1. Read the question carefully and identify the main aspects that must be covered for a complete response.

2. Read the answer and compare it to the question. Check if the answer addresses all essential aspects, main ideas, and pertinent details required by the question.

3. Assign a score for completeness based on the Evaluation Criteria.

Example:

Question:
{}

Answer:
{}

Completeness Evaluation Form (return numeric scores ONLY):

\end{tcolorbox}
}

\end{document}